\documentclass[runningheads]{llncs}
\usepackage[T1]{fontenc}
\usepackage{booktabs}
\usepackage{url}
\usepackage{amsmath}
\usepackage{subcaption}
\usepackage{verbatim}

\usepackage{graphicx}
\begin{document}
\title{MiNER: A Two-Stage Pipeline for Metadata Extraction from Municipal Meeting Minutes}

\titlerunning{MiNER: Metadata Extraction from Municipal Minutes}

%
%
\author{
Rodrigo~Batista\inst{1,2}\orcidID{0009-0004-0431-6042} \and
Luís~Filipe~Cunha\inst{1,2}\orcidID{0000-0003-1365-0080} \and
Purificação~Silvano\inst{1,3}\orcidID{0000-0001-8057-5338} \and
Nuno~Guimarães\inst{1,2}\orcidID{0000-0003-2854-2891} \and
Alípio~Jorge\inst{1,2}\orcidID{0000-0002-5475-1382} \and
Evelin~Amorim\inst{1,2}\orcidID{0000-0003-1343-939X} \and
Ricardo~Campos\inst{1,4}\orcidID{0000-0002-8767-8126}
}
\authorrunning{R. Batista et al.}
%
\institute{
INESC TEC, Porto, Portugal 
\email{\{luis.f.cunha, rodrigo.f.batista, evelin.f.amorim, alipio.jorge, nuno.r.guimaraes, purificacao.silvano\}@inesctec.pt}\\
\and
Faculdade de Ciências, Universidade do Porto, Porto, Portugal
\and
Faculdade de Letras, Universidade do Porto, Porto, Portugal
\and
University of Beira Interior, Covilhã, Portugal \\
\email{\{ricardo.campos\}@ubi.pt}
}
\maketitle              

\setcounter{footnote}{0}

\begin{abstract}
Municipal meeting minutes are official documents of local governance, exhibiting heterogeneous formats and writing styles. Effective information retrieval (IR) requires identifying metadata such as meeting number, date, location, participants, and start/end times, elements that are rarely standardized or easy to extract automatically. Existing named entity recognition (NER) models are ill-suited to this task, as they are not adapted to such domain-specific categories. In this paper, we propose a two-stage pipeline for metadata extraction from municipal minutes. First, a question answering (QA) model identifies the opening and closing text segments containing metadata. Transformer-based models (BERTimbau and XLM-RoBERTa with and without a CRF layer) are then applied for fine-grained entity extraction and enhanced through deslexicalization. To evaluate our proposed pipeline, we benchmark both open-weight (Phi) and closed-weight (Gemini) LLMs, assessing predictive performance, inference cost, and carbon footprint. Our results demonstrate strong in-domain performance, better than larger general-purpose LLMs. However, cross-municipality evaluation reveals reduced generalization reflecting the variability and linguistic complexity of municipal records. 
This work establishes the first benchmark for metadata extraction from municipal meeting minutes, providing a solid foundation for future research in this domain.
\end{abstract}
\section{Introduction}

Municipal meeting minutes are important documents of local governance, providing official records of decisions, participants, and voting outcomes. Beyond their legal, administrative, and civic importance, these documents are increasingly used for research \cite{councilproject,10.1007/978-3-031-56063-7_17}. Effective information retrieval (IR) from these records requires extracting structured metadata, including meeting number, type of meeting, date, location, participants, and the session times, which are consistently present in parliamentary datasets\cite{erjavec2023parlamint,hyvonen2024parliamentsampo} and meeting minutes. While pre-trained NER models perform well on general-purpose categories (e.g, person, organization, location), they fail to capture the fine-grained, domain-specific metadata typical of these records. Dedicated models for this domain remain, to the best of our knowledge, nonexistent. Therefore, we propose a two-stage pipeline for automatic metadata extraction. First, a question answering (QA) model identifies the opening and closing segments of each document, reducing noise and focusing subsequent processing. Second, transformer-based models perform fine-grained entity recognition, enhanced via deslexicalization to improve generalization across municipalities. We evaluate our approach on a new dataset of 120 Portuguese municipal minutes, comprising a total of 180 metadata segments, with an English automatic translation for cross-lingual testing. Comparisons with open (Phi~\cite{abdin2024phi}) and closed (Gemini~\cite{team2024gemini}) weights LLMs regarding predictive performance, inference time and estimated carbon footprint show that specialized transformer models, outperform general-purpose LLMs in every aspects, while revealing the challenges of cross-municipality generalization. The main contribution are 1) a two-stage pipeline, combining QA-based segmentation with transformer-based NER for fine-grained entity recognition and 2) the release of the corresponding dataset\footnote{\url{https://github.com/LIAAD/MiNER}}, and fine-tuned models\footnote{\url{https://huggingface.co/collections/liaad/citilink}}, establishing a benchmark for metadata extraction from municipal minutes.
\section{Related Work}

Metadata extraction has been explored across various domains, being crucial for indexing, retrieval, and structured analysis in complex documents. Resources such as IsraParlTweet~\cite{mor-lan-etal-2024-israparltweet} and Watanabe et al.~\cite{watanabe2024llmmetadata} demonstrate the utility of metadata in parliamentary and scholarly contexts, while also highlighting the limitations of LLMs for automatic metadata extraction. However, most approaches assume metadata sections are already identifiable, overlooking the challenge of locating them automatically. This problem can instead be framed as a boundary detection task, for which QA-based formulations ~\cite{rajpurkar2016squad100000questionsmachine,rajpurkar2018knowdontknowunanswerable} have proven effective in isolating relevant spans
 and identifying the start and end points of text segments. Following boundary identification, NER can be applied to extract fine-grained entities such as dates, locations, participants, and other administrative details~\cite{spinosa2009nlp,sleimi2021automated}. Our work differs from prior research by focusing on municipal meeting minutes, a domain that exhibits greater stylistic variability and finer-grained metadata categories than parliamentary or scholarly texts. Furthermore, unlike previous approaches that assume metadata is pre-segmented, we explicitly identify metadata segments before entity extraction. This combination of domain-specific segmentation and fine-grained NER makes our approach distinct within the broader landscape of metadata extraction.

\section{MiNER: Minutes Named Entity Recognizer Pipeline}

Our pipeline has two main phases, transforming a raw document $D$ into a set of labeled metadata entities extracted from the specific document region. Formally, let $D = \{s_1, s_2, \dots, s_n\}$
be an ordered sequence of sentences, and let $\mathcal{M} = \{m_1,\dots,m_K\}$ denote the set of metadata categories of interest (e.g., meeting number and type, date, location, participants, session times).
We decompose the task into two stages as follows.


\vspace{0.5em}

\noindent\textbf{Stage 1 - Metadata Boundary Detection (MBD)}. We start by identifying the opening and closing sections of the minutes where metadata is found. This is approached using extractive QA. For each document $D$, we employ two prompts $p_t$,  one for each segment $t \in \{\text{opening}, \text{closing}\}$, which output the positions $\hat{a}_t$ and $\hat{b}_t$ of the last and the first sentences of the opening and closing segment, respectively. 
The identified segments are then defined as
$
S_t = \{ s_{\hat{a}_t}, \dots, s_{\hat{b}_t} \}, \quad t \in \{\text{opening}, \text{closing}\},
$
and the concatenation $R = S_{\text{opening}} \cup S_{\text{closing}}$ is the reduced text region passed to Stage 2.
Formally, the Stage 1 model is defined as
\[
f_{\text{BD}}(D, p_t; \theta_{\text{QA}}) \rightarrow R,
\]

\noindent where $\theta_{\text{QA}}$ is the learned parameters of the QA model. The QA approach follows a SQuAD v2-style setup~\cite{rajpurkar2018knowdontknowunanswerable,rajpurkar2016squad100000questionsmachine}. The model outputs either a text span $(\hat{a}_t, \hat{b}_t)$ or a null prediction, corresponding to the empty segment.  

\vspace{1em}
\noindent\textbf{Stage 2 - Metadata Entity Recognition (MER)}. 
Given the reduced text $R$, the goal is to identify and classify all metadata mentions. Formally,
\[
f_{\text{NER}}(R; \theta_{\text{NER}}) \rightarrow \mathcal{E},
\]
where $\mathcal{E} = \{ e_1, \dots, e_R \}$ is the set of extracted entities, each defined by its type $\tau \in \mathcal{M}$ and a consecutive token span $(u,v)$ within $R$, and $\theta_{\text{NER}}$ represents the learned parameters of the NER model. 
The model is trained using standard token-level classification objectives (BIO or CRF), restricted to $R$ to reduce noise and improve precision.
To promote cross-municipality generalization, a deslexicalization function $\phi(\cdot)$ is applied prior to training 
\cite{sousa2025enhancingportuguesevarietyidentification}. In this context, participants and location names are replaced with synthetic values (60\% probability) using \texttt{Faker} library \cite{faker} and dates and times are randomly varied in format or content (30\% probability). Municipality mentions are also replaced with a placeholder token (\texttt{@MUNICIPIO}). 
The pipeline is composed of the two stages:
\[
\mathcal{F}(D; \theta_{\text{QA}}, \theta_{\text{NER}}) 
= f_{\text{NER}}\big(f_{\text{BD}}(D; \theta_{\text{QA}}); \theta_{\text{NER}}\big) 
\rightarrow \mathcal{E},
\]


\section{Experimental Setup}

To evaluate our pipeline, we conducted experiments covering both stages.

\noindent\textbf{Evaluation Metrics}. For MBD, we report Exact Match (EM), which gives the proportion of predictions that exactly match the ground-truth span, and F1-score, which measures token-level overlap between predicted and ground truth spans~\cite{rajpurkar2018knowdontknowunanswerable,rajpurkar2016squad100000questionsmachine}. For MER, we report precision, recall, and F1-score. In experiments involving LLMs, we also measure inference cost and estimated carbon footprint, using \texttt{Code Carbon} library \cite{codecarbon}. In addition, cross-municipality generalization was evaluated using a leave-one-out setup, training on five municipalities and testing on the remaining one. We also simulated real deployment with scarce data adding one annotated minute at a time to assess the number of annotated documents required to achieve stable performance. 

\noindent\textbf{Dataset}. We make use of the newly introduced CitiLink dataset \cite{citilink2025dataset}, which consists of 120 Portuguese municipal meeting minutes from six municipalities, manually annotated with multiple layers of information \cite{citilink-dataset}, with the annotation guidelines publicly available in the repository. In this work, we focus exclusively on the metadata annotations, consisting of 32,364 metadata segments. Annotated entities cover \textit{meeting number} (120), \textit{date} (120), \textit{location} (100), \textit{start time} (120), \textit{end time} (100), \textit{meeting type} (101), and \textit{participants} (president (119) and councilors (966), including presence status), totaling eight distinct categories. Personal information was anonymized (e.g., replaced with ``\texttt{***}'') to ensure privacy and compliance with data handling regulations. For cross-lingual evaluation, an English version was automatically translated using Azure Translator, and annotations were aligned with LinguAligner \cite{lingualigner}. The test set in English was further reviewed and corrected by human annotators. The dataset was subsequently adapted for the two stages of our pipeline. For MBD, each document was converted into SQuAD v2-style format, where the full text serves as context, and each prompt (opening or closing segment) is paired with the corresponding answer span. For MER, only the opening and closing segments from Stage~1 were considered. Due to space limitations, additional details are described in the repository.

\noindent\textbf{Models and Baselines}. For MBD, we used \textbf{deepset/xlm-roberta-large-squad2}~\cite{deepset-xlm-roberta-large-squad2}. Training ran for 3 epochs (lr = 3e-5, batch = 8, weight decay = 0.01, max length = 512, stride = 128). We compared against unsupervised baselines: \textbf{BM25}, \cite{bm25} which ranks sentences by lexical relevance, and a \textbf{Dense Retriever} \cite{karpukhin-etal-2020-dense} using semantic embeddings for similarity search.
For MER, we resorted to \textbf{BERTimbau-large}~\cite{10.1007/978-3-030-61377-8_28} for Portuguese, and \textbf{XLM-RoBERTa-large}~\cite{xlm-roberta} for English, with and without a CRF layer. Training was conducted for 15 epochs with early stopping (patience of 3 epochs), learning rate 2e-5, batch size 2 per device with gradient accumulation over 4 steps, and weight decay 0.01. We applied a global document-level split: 60\% training, 20\% validation, and 20\% testing, maintaining diversity across municipalities and entity types, and avoiding overlap between minutes.
Finally, these approaches were compared with Gemini and Phi utilizing the LangExtract library \cite{langextract}. More details such as the prompt and hardware used are available in the repository.

\section{Results}

\textbf{MBD}. XLM-RoBERTa clearly outperforms BM25 and dense retrievers across all metrics and languages (Table \ref{tab:qaresults}), with a slight performance drop in English likely due to translation noise and cross-lingual transfer limitations. 

\begin{table}[h]
    \centering
    \caption{Results on the QA (Portuguese and English).}
    \label{tab:qaresults}
    \renewcommand{\arraystretch}{1.0}
    \begin{tabular}{l|cc|cc} 
        \toprule
        \textbf{Model} & \multicolumn{2}{c|}{\textbf{F1-score}} & \multicolumn{2}{c|}{\textbf{EM}} \\
         & \textbf{pt} & \textbf{en} & \textbf{pt} & \textbf{en} \\ 
        \midrule
        BM25 & 0.094 & 0.093 & 0.0 & 0.021  \\
        Dense Retriever & 0.077 & 0.111 & 0.0 & 0.083 \\
        \textbf{XLM-Roberta} & \textbf{0.826} & \textbf{0.714} & \textbf{0.792} & \textbf{0.604} \\
        \bottomrule
    \end{tabular}
\end{table}

\noindent\textbf{MER} Table \ref{tab:globalresults} reports the MER results on the global split. All transformer-based models achieve strong performance, with BERTimbau obtaining the best results in Portuguese (\textbf{0.96}) and XLM-RoBERTa+Deslex in English (\textbf{0.94}). Entity-level results are not tabulated, but are available in our repository. Overall, the models perform particularly well on core metadata categories such as date, start/end time, and meeting number, which are extracted with near-perfect accuracy (F1 = 1.00). A detailed error analysis shows that most remaining errors are boundary-related, indicating that the models generally identify the correct entity type but struggle with precise span delimitation. In contrast, participant-related entities exhibit higher variability: while “councilors present” reaches a F1-score of 0.98, absent and replaced members range between 0.28 and 0.70, with frequent confusion between presence states (present, absent, and substituted). Meeting type categories are similarly imbalanced, with “ordinary meetings” being robustly detected (F1 = 0.95–1.0), whereas “extraordinary meetings” underperform due to the very limited number of examples (only six instances).

\begin{table}[b]
    \centering
    \caption{Results on the global split (Portuguese and English).}
    \label{tab:globalresults}
    \renewcommand{\arraystretch}{1.0}
    \setlength{\tabcolsep}{3pt} 
    \begin{tabular}{l|cc|cc|cc} 
        \toprule
        \textbf{Model} & \multicolumn{2}{c|}{\textbf{F1-score}} & \multicolumn{2}{c|}{\textbf{Precision}} & \multicolumn{2}{c}{\textbf{Recall}} \\
         & \textbf{pt} & \textbf{en} & \textbf{pt} & \textbf{en} & \textbf{pt} & \textbf{en} \\ 
        \midrule
        \textbf{BERTimbau} & \textbf{0.96} & 0.94 & \textbf{0.96} & 0.94 & \textbf{0.96} & 0.94 \\
        XLM-RoBERTa & 0.93 & 0.94 & 0.94 & 0.94 & 0.92 & 0.93 \\
        BERTimbau + CRF & 0.95 & 0.93 & 0.95 & 0.92 & 0.96 & 0.95 \\
        XLM-RoBERTa + CRF & 0.90 & 0.89 & 0.88 & 0.88 & 0.93 & 0.91 \\
        BERTimbau + Deslex & 0.96 & 0.93 & 0.96 & 0.93 & 0.96 & 0.94 \\
        \textbf{XLM-RoBERTa + Deslex} & 0.96 & \textbf{0.94} & 0.96 & \textbf{0.94} & 0.97 & \textbf{0.94} \\ 
        \bottomrule
    \end{tabular}
\end{table}

Leave-one-out results (Table \ref{tab:leaveoneoutresults}) show reduced generalization compared to the global split (F1 = 0.80 in Portuguese and F1 = 0.72 in English). Detailed error analysis reveals that this degradation is primarily driven by an increase in boundary errors, which become more frequent with deslexicalization in both languages. In Portuguese, the performance drop is mainly reflected in lower recall, indicating missed entities after the removal of lexical cues, whereas in English, the decrease is largely associated with overgeneralization and less precise span boundaries. In incremental evaluation (Table \ref{tab:incrementalresults}), performance improves sharply after the first annotated minutes, with most municipalities reaching F1-scores above 0.95 by the third iteration, showing that minimal local supervision enables rapid model specialization without full retraining.

\begin{table}[t]
\centering
\caption{Leave-one-out results for the Portuguese  (left) and English (right) models tested without (Base) and with (Deslex) deslexicalization.}
\label{tab:leaveoneoutresults}
\renewcommand{\arraystretch}{1.0}

\begin{subtable}[t]{0.45\linewidth}
\centering
\begin{tabular}{l|ccc}
\toprule
\textbf{BERTimbau} & \textbf{F1} & \textbf{Precision} & \textbf{Recall} \\
\midrule
Base & 0.80 & 0.79 & 0.81 \\
Deslex & 0.78 & 0.79 & 0.79 \\
\bottomrule
\end{tabular}
\end{subtable}\hfill
\begin{subtable}[t]{0.48\linewidth}
\centering
\begin{tabular}{c|ccc}
\toprule
\textbf{XML-Roberta} & \textbf{F1} & \textbf{Precision} & \textbf{Recall} \\
\midrule
Base & 0.72 & 0.73 & 0.72 \\
Deslex & 0.70 & 0.70 & 0.72 \\
\bottomrule
\end{tabular}
\end{subtable}
\end{table}

\begin{table}[t]
    \centering
    \caption{Incremental evaluation results per municipality (F1-score).}
    \label{tab:incrementalresults}
    \renewcommand{\arraystretch}{1.0}
    \setlength{\tabcolsep}{5pt}
    \resizebox{\columnwidth}{!}{%
    \begin{tabular}{l|cccccc}
        \toprule
        \textbf{Examples} & \textbf{Alandroal} & \textbf{Campo Maior} & \textbf{Covilhã} & \textbf{Fundão} & \textbf{Guimarães} & \textbf{Porto} \\
        \midrule
        0 & 0.662 & 0.758 & 0.802 & 0.899 & 0.784 & 0.820 \\
        1 & 0.960 & 0.995 & 0.872 & 0.980 & 0.915 & 0.971 \\
        2 & 0.977 & 0.995 & 0.923 & 0.986 & 0.917 & 0.999 \\
        3 & 0.995 & 0.995 & 0.929 & 0.989 & 0.941 & 0.989 \\
        4 & 0.995 & 0.995 & 0.953 & 0.989 & 0.949 & 0.993 \\
        5 & 0.995 & 0.995 & 0.947 & 0.993 & 0.963 & 0.996 \\
        \bottomrule
    \end{tabular}
    }
\end{table}

 We also compare our pipeline with general-purpose LLMs. While LLM comparison is included, our primary goal is to evaluate the MiNER pipeline rather to exhaustively benchmark LLMs. Thus, we focus on open-weight (Phi) and freely accessible (Gemini) models to ensure reproducibility and enable practitioners with limited budgets to replicate our findings. The \textbf{Gemini} model achieved a modest F1-score (0.27), characterized by high precision (0.83) but low recall (0.16), indicating that, while it correctly identifies some entities, it often fails to capture their structure in municipal minutes, leading to fragmented outputs. The \textbf{Phi} model could not be reliably evaluated because of inconsistent JSON outputs, though structured decoding techniques could mitigate this limitation. In contrast, the pipeline (MBD + NER) consistently produced fully structured outputs, achieving substantially higher performance while requiring over \textbf{1,800 times less inference time} (approximately 0.4\,s vs.\ 737\,s) and emitting nearly \textbf{400 times lower carbon emissions} ($6\times10^{-6}$ vs.\ $2.2\times10^{-3}$\,kg\,CO$_2$e). These results highlight that, despite their flexibility, the evaluated LLMs are less suited than task-specific models for structured metadata extraction in this domain.

Finally, we conducted an ablation experiment to assess the impact of the first stage of the pipeline. When the NER component was applied directly to the full documents, without the first stage, the F1-score slightly decreased from \textbf{0.965} to \textbf{0.945}. Although this difference is relatively small, the MBD stage plays a crucial role in practice by substantially reducing the amount of text processed by the NER model, decreasing the training time from approximately \textbf{50 minutes} to only \textbf{1 minute}. Moreover, by restricting NER to semantically relevant regions in such documents, the segmentation step reduces noise, and enables the reuse of the extracted segments for other downstream tasks that rely on accurate document boundaries, such as meeting summarization, text segmentation, or voting identification.

\section{Conclusion and Future Work}

In this paper, we present MiNER, a two-stage pipeline for metadata extraction from municipal meeting minutes, which combines QA-based boundary detection with transformer-based NER models. MiNER achieves strong results across Portuguese and English datasets, demonstrating that domain-specific fine-tuning can outperform large general-purpose LLMs while remaining computationally efficient. Despite its strengths, MiNER’s generalization across municipalities remains limited, as stylistic and structural variability hinders transfer learning. Moreover, the segmentation stage, although beneficial for noise reduction, offers only marginal gains in end-to-end accuracy. Future work will explore adaptive fine-tuning strategies, domain-agnostic representations, and integration with summarization or document-linking pipelines to extend MiNER’s applicability to broader information management tasks.

\begin{credits}
\subsubsection*{Preprint and Version of Record}
This preprint has not undergone peer review (when applicable) or any post-submission improvements or corrections. The Version of Record of this contribution is published in {Advances in Information Retrieval. ECIR 2026. Lecture Notes in Computer Science, vol 16484. Springer, Cham}, and is available online at https://doi.org/10.1007/978-3-032-21300-6\_33.

\subsubsection*{\ackname}
This work is financed by National Funds through the FCT - Fundação para a Ciência e a Tecnologia, I.P. (Portuguese Foundation for Science and Technology) within the project CitiLink, with reference  2024.07509.IACDC/2024 (DOI \url{https://doi.org/10.54499/2024.07509.IACDC}) within the scope of PRR funding, investment “RE-C05-i08 - More Digital Science”.
\end{credits}
\bibliographystyle{splncs04}
\bibliography{mybib}

@inproceedings{mor-lan-etal-2024-israparltweet,
    title = "{I}sra{P}arl{T}weet: The Israeli Parliamentary and {T}witter Resource",
    author = "Mor-Lan, Guy  and
      Levi, Effi  and
      Sheafer, Tamir  and
      Shenhav, Shaul R.",
    editor = "Calzolari, Nicoletta  and
      Kan, Min-Yen  and
      Hoste, Veronique  and
      Lenci, Alessandro  and
      Sakti, Sakriani  and
      Xue, Nianwen",
    booktitle = "Proceedings of the 2024 Joint International Conference on Computational Linguistics, Language Resources and Evaluation (LREC-COLING 2024)",
    month = may,
    year = "2024",
    address = "Torino, Italia",
    publisher = "ELRA and ICCL",
    url = "https://aclanthology.org/2024.lrec-main.819/",
    pages = "9372--9381"
}

@article{sleimi2021automated,
  title={An automated framework for the extraction of semantic legal metadata from legal texts},
  author={Sleimi, Amin and Sannier, Nicolas and Sabetzadeh, Mehrdad and Briand, Lionel and Ceci, Marcello and Dann, John},
  journal={Empirical Software Engineering},
  volume={26},
  number={3},
  pages={43},
  year={2021},
  publisher={Springer}
}

@inproceedings{spinosa2009nlp,
  title={NLP-based metadata extraction for legal text consolidation},
  author={Spinosa, PierLuigi and Giardiello, Gerardo and Cherubini, Manola and Marchi, Simone and Venturi, Giulia and Montemagni, Simonetta},
  booktitle={Proceedings of the 12th international conference on artificial intelligence and law},
  pages={40--49},
  year={2009}
}

@InProceedings{10.1007/978-3-030-61377-8_28,
author="Souza, F{\'a}bio
and Nogueira, Rodrigo
and Lotufo, Roberto",
editor="Cerri, Ricardo
and Prati, Ronaldo C.",
title="BERTimbau: Pretrained BERT Models for Brazilian Portuguese",
booktitle="Intelligent Systems",
year="2020",
publisher="Springer International Publishing",
address="Cham",
pages="403--417",
isbn="978-3-030-61377-8"
}

@misc{rajpurkar2016squad100000questionsmachine,
      title={SQuAD: 100,000+ Questions for Machine Comprehension of Text}, 
      author={Pranav Rajpurkar and Jian Zhang and Konstantin Lopyrev and Percy Liang},
      year={2016},
      eprint={1606.05250},
      archivePrefix={arXiv},
      primaryClass={cs.CL},
      url={https://arxiv.org/abs/1606.05250}, 
}

@misc{rajpurkar2018knowdontknowunanswerable,
      title={Know What You Don't Know: Unanswerable Questions for SQuAD}, 
      author={Pranav Rajpurkar and Robin Jia and Percy Liang},
      year={2018},
      eprint={1806.03822},
      archivePrefix={arXiv},
      primaryClass={cs.CL},
      url={https://arxiv.org/abs/1806.03822}, 
}

@article{team2024gemini,
  title={Gemini 1.5: Unlocking multimodal understanding across millions of tokens of context},
  author={Team, Gemini and Georgiev, Petko and Lei, Ving Ian and Burnell, Ryan and Bai, Libin and Gulati, Anmol and Tanzer, Garrett and Vincent, Damien and Pan, Zhufeng and Wang, Shibo and others},
  journal={arXiv preprint arXiv:2403.05530},
  year={2024}
}

@article{abdin2024phi,
  title={Phi-4 technical report},
  author={Abdin, Marah and Aneja, Jyoti and Behl, Harkirat and Bubeck, S{\'e}bastien and Eldan, Ronen and Gunasekar, Suriya and Harrison, Michael and Hewett, Russell J and Javaheripi, Mojan and Kauffmann, Piero and others},
  journal={arXiv preprint arXiv:2412.08905},
  year={2024}
}

@misc{councilproject,
  author       = {{Council Data Project}},
  title        = {councilproject},
  howpublished = {\url{https://councildataproject.org/}},
  note         = {Accessed: 2025-10-10}
}

@misc{deepset-xlm-roberta-large-squad2,
  author       = {{deepset}},
  title        = {xlm-roberta-large-squad2},
  howpublished = {\url{https://huggingface.co/deepset/xlm-roberta-large-squad2}},
  note         = {Accessed: 2025-10-11}
}

@misc{faker,
  author       = {{Faker Contributors}},
  title        = {Faker},
  howpublished = {\url{https://pypi.org/project/Faker/}},
  note         = {Accessed: 2025-09-15}
}

@misc{xlm-roberta,
  author       = {{Hugging Face}},
  title        = {xlm-roberta},
  howpublished = {\url{https://huggingface.co/docs/transformers/model_doc/xlm-roberta
}},
  note         = {Accessed: 2025-09-15}
}

@misc{codecarbon,
  author       = {{CodeCarbon Team}},
  title        = {CodeCarbon},
  howpublished = {\url{https://codecarbon.io/}},
  note         = {Accessed: 2025-10-01}
}

@misc{langextract,
  author       = {Goel, Akshay},
  year         = {2025},
  title        = {LangExtract (Version 1.1.1) [Computer software]},
  doi          = {10.5281/zenodo.17015089},
  url          = {https://doi.org/10.5281/zenodo.17015089}
}

@misc{lingualigner,
  author       = {{LinguAligner Developers}},
  title        = {LinguAligner},
  howpublished = {\url{https://pypi.org/project/LinguAligner/}},
  note         = {Accessed: 2025-09-15}
}

@inproceedings{karpukhin-etal-2020-dense,
    title = "Dense Passage Retrieval for Open-Domain Question Answering",
    author = "Karpukhin, Vladimir  and
      Oguz, Barlas  and
      Min, Sewon  and
      Lewis, Patrick  and
      Wu, Ledell  and
      Edunov, Sergey  and
      Chen, Danqi  and
      Yih, Wen-tau",
    editor = "Webber, Bonnie  and
      Cohn, Trevor  and
      He, Yulan  and
      Liu, Yang",
    booktitle = "Proceedings of the 2020 Conference on Empirical Methods in Natural Language Processing (EMNLP)",
    month = nov,
    year = "2020",
    address = "Online",
    publisher = "Association for Computational Linguistics",
    url = "https://aclanthology.org/2020.emnlp-main.550/",
    doi = "10.18653/v1/2020.emnlp-main.550",
    pages = "6769--6781"
}

@article{bm25,
author = {Robertson, Stephen and Zaragoza, Hugo},
year = {2009},
month = {01},
pages = {333-389},
title = {The Probabilistic Relevance Framework: BM25 and Beyond},
volume = {3},
journal = {Foundations and Trends in Information Retrieval},
doi = {10.1561/1500000019}
}

@misc{sousa2025enhancingportuguesevarietyidentification,
      title={Enhancing Portuguese Variety Identification with Cross-Domain Approaches}, 
      author={Hugo Sousa and Rúben Almeida and Purificação Silvano and Inês Cantante and Ricardo Campos and Alípio Jorge},
      year={2025},
      eprint={2502.14394},
      archivePrefix={arXiv},
      primaryClass={cs.CL},
      url={https://arxiv.org/abs/2502.14394}, 
}

@InProceedings{watanabe2024llmmetadata,
author="Watanabe, Yu
and Ito, Koichiro
and Matsubara, Shigeki",
editor="Oliver, Gillian
and Frings-Hessami, Viviane
and Du, Jia Tina
and Tezuka, Taro",
title="Capabilities and Challenges of LLMs in Metadata Extraction from Scholarly Papers",
booktitle="Sustainability and Empowerment in the Context of Digital Libraries",
year="2025",
publisher="Springer Nature Singapore",
address="Singapore",
pages="280--287",
isbn="978-981-96-0865-2"
}

@article{hyvonen2024parliamentsampo,
author = {Eero Hyvönen and Laura Sinikallio and Petri Leskinen and Senka Drobac and Rafael Leal and Matti La Mela and Jouni Tuominen and Henna Poikkimäki and Heikki Rantala},
title ={Publishing and using parliamentary Linked Data on the Semantic Web: ParliamentSampo system for Parliament of Finland},
journal = {Semantic Web},
volume = {16},
number = {1},
pages = {SW-243683},
year = {2025},
doi = {10.3233/SW-243683},
URL = {   
        https://journals.sagepub.com/doi/abs/10.3233/SW-243683  
},
eprint = { 
        https://journals.sagepub.com/doi/pdf/10.3233/SW-243683
}
}

@InProceedings{10.1007/978-3-031-56063-7_17,
author="Jain, Arihant
and Sharma, Raksha",
editor="Goharian, Nazli
and Tonellotto, Nicola
and He, Yulan
and Lipani, Aldo
and McDonald, Graham
and Macdonald, Craig
and Ounis, Iadh",
title="Enhancing Legal Named Entity Recognition Using RoBERTa-GCN with CRF: A Nuanced Approach for Fine-Grained Entity Recognition",
booktitle="Advances in Information Retrieval",
year="2024",
publisher="Springer Nature Switzerland",
address="Cham",
pages="261--267",
isbn="978-3-031-56063-7"
}

@article{erjavec2023parlamint,
  author    = {Tomaž Erjavec and Maciej Ogrodniczuk and Petya Osenova and others},
  title     = {The ParlaMint corpora of parliamentary proceedings},
  journal   = {Language Resources and Evaluation},
  volume    = {57},
  pages     = {415--448},
  year      = {2023},
  doi       = {10.1007/s10579-021-09574-0},
  url       = {https://doi.org/10.1007/s10579-021-09574-0}
}

@InProceedings{citilink-dataset,
author="Campos, Ricardo
and Pacheco, Ana Filipa
and Fernandes, Ana Luísa
and Cantante, Inês
and Rebouças, Rute
and Cunha, Luís Filipe
and Isidro, José
and Evans, José Pedro
and Marques, Miguel
and Batista, Rodrigo
and Amorim, Evelin
and Jorge, Alípio
and Guimarães, Nuno
and Nunes, Sérgio
and Leal, António
and Silvano, Purificação",
editor="Anand, Avishek
and Ren, Zhaochun
and Verberne, Suzan
and Jatowt, Adam
and Campos, Ricardo
and Lan, Yanyan
and MacAvaney, Sean
and Aliannejadi, Aliannejadi
and Bauer, Christine
and Mansoury, Masoud",
title="CitiLink-Minutes: A Multilayer Annotated Dataset of Municipal Meeting Minutes",
booktitle="Advances in Information Retrieval",
year="2026",
publisher="Springer Nature Switzerland",
address="Cham",
pages=""
}

@misc{citilink2025dataset,
  author       = {Ricardo Campos and Ana Filipa Pacheco and Ana Luísa Fernandes and Inês Cantante and Rute Rebouças and Luís Filipe Cunha and José Isidro and José Evans and Miguel Marques and Rodrigo Batista and Evelin Amorim and Alípio Jorge and Nuno Guimarães and Sérgio Nunes and António Leal and Purificação Silvano},
  title        = {{CitiLink-Minutes: A Multilayer Annotated Dataset of Municipal Meeting Minutes}},
  year         = {2025},
  doi          = {10.25747/7KG6-1K22},
  institution  = {INESC TEC}
}
\end{document}